\relax

\documentclass[11pt,a4paper]{article}
\usepackage[hyperref]{naaclhlt2018}
\usepackage{times}
\usepackage{latexsym}
\usepackage{url}
\usepackage{graphicx}
\usepackage[ruled,linesnumbered]{algorithm2e}
\usepackage{amsmath}
\usepackage{amssymb}
\usepackage{cuted}
\usepackage{subfig}
\aclfinalcopy
 
\title{\textsc{kbgan}: Adversarial Learning for Knowledge Graph Embeddings}
\author{Liwei Cai \\
  Department of Electronic Engineering \\
  Tsinghua University \\
  Beijing 100084 China \\
  {\tt cai.lw123@gmail.com} \\\And
  William Yang Wang \\
  Department of Computer Science \\
  University of California, Santa Barbara \\
  Santa Barbara, CA 93106 USA\\
  {\tt william@cs.ucsb.edu} \\}

\begin{document}
\maketitle
\begin{abstract}
We introduce \textsc{kbgan}, an adversarial learning framework to improve the performances of a wide range of existing knowledge graph embedding models. Because knowledge graphs typically only contain positive facts, sampling useful negative training examples is a non-trivial task. Replacing the head or tail entity of a fact with a uniformly randomly selected entity is a conventional method for generating negative facts, but the majority of the generated negative facts can be easily discriminated from positive facts, and will contribute little towards the training. Inspired by generative adversarial networks (\textsc{gan}s), we use one knowledge graph embedding model as a negative sample generator to assist the training of our desired model, which acts as the discriminator in \textsc{gan}s. 
This framework is independent of the concrete form of generator and discriminator, and therefore can utilize a wide variety of knowledge graph embedding models as its building blocks. In experiments, we adversarially train two translation-based models, \textsc{TransE} and \textsc{TransD}, each with assistance from one of the two probability-based models, \textsc{DistMult} and \textsc{ComplEx}. We evaluate the performances of \textsc{kbgan} on the link prediction task, using three knowledge base completion datasets: FB15k-237, WN18 and WN18RR. Experimental results show that adversarial training substantially improves the performances of target embedding models under various settings.
\end{abstract}

\section{Introduction}
Knowledge graph~\cite{dong2014knowledge} is a powerful graph structure that can provide direct access of knowledge to users via various applications such as structured search, question answering, and intelligent virtual assistant. A common representation of knowledge graph beliefs is in the form of a discrete relational triple such as \emph{LocatedIn(NewOrleans,Louisiana)}.

A main challenge for using discrete representation of knowledge graph is the lack of capability of accessing the similarities among different entities and relations. Knowledge graph embedding (KGE) techniques (e.g., \textsc{rescal} \cite{nickel2011three}, \textsc{TransE} \cite{bordes2013translating}, \textsc{DistMult} \cite{yang2015embedding}, and \textsc{ComplEx} \cite{trouillon2016complex}) have been proposed in recent years to deal with the issue. The main idea is to represent the entities and relations in a vector space, and one can use machine learning technique to learn the continuous representation of the knowledge graph in the latent space.

However, even steady progress has been made in developing novel algorithms for knowledge graph embedding, there is still a common challenge in this line of research. For space efficiency, common knowledge graphs such as Freebase~\cite{bollacker2008freebase}, Yago~\cite{suchanek2007yago}, and NELL~\cite{mitchell2015never} by default only stores beliefs, rather than disbeliefs. Therefore, when training the embedding models, there is only the natural presence of the positive examples. To use negative examples, a common method is to remove the correct tail entity, and randomly sample from a uniform distribution~\cite{bordes2013translating}. Unfortunately, this approach is not ideal, because the sampled entity could be completely unrelated to the head and the target relation, and thus the quality of randomly generated negative examples is often poor (e.g, \emph{LocatedIn(NewOrleans,BarackObama)}). Other approach might leverage external ontological constraints such as entity types \cite{krompass2015type} to generate negative examples, but such resource does not always exist or accessible.

\begin{table*}[t]
\centering
\small
\begin{tabular}{|l|l|l|l|}
\hline
\textbf{Model} & \textbf{Score function $f(h, r, t)$} & \textbf{Number of parameters} \\ \hline
\textsc{TransE} & $||\mathbf{h}+\mathbf{r}-\mathbf{t}||_{1/2}$ & $k|\mathcal{E}|+k|\mathcal{R}|$ \\ \hline
\textsc{TransD} & $||(\mathbf{I}+\mathbf{r_p}\mathbf{h_p}^T)\mathbf{h}+\mathbf{r}-(\mathbf{I}+\mathbf{r_p}\mathbf{t_p}^T)\mathbf{t}||_{1/2}$ & $2k|\mathcal{E}|+2k|\mathcal{R}|$ \\ \hline
\textsc{DistMult} & $<\mathbf{h}, \mathbf{r}, \mathbf{t}>$ ($=\sum_{i=1}^{k} h_i r_i t_i$) & $k|\mathcal{E}|+k|\mathcal{R}|$ \\ \hline
\textsc{ComplEx} & $<\mathbf{h}, \mathbf{r}, \mathbf{\bar{t}}>$ ($\mathbf{h}, \mathbf{r}, \mathbf{t} \in \mathbb{C}^k$) & $2k|\mathcal{E}|+2k|\mathcal{R}|$ \\ \hline\hline
\textsc{TransH} & $||(\mathbf{I}-\mathbf{r_p}\mathbf{r_p}^T)\mathbf{h}+\mathbf{r}-(\mathbf{I}+\mathbf{r_p}\mathbf{r_p}^T)\mathbf{t}||_{1/2}$ & $k|\mathcal{E}|+2k|\mathcal{R}|$ \\ \hline
\textsc{TransR} & $||\mathbf{W_r}\mathbf{h}+\mathbf{r}-\mathbf{W_r}\mathbf{t}||_{1/2}$ & $k|\mathcal{E}|+(k^2+k)|\mathcal{R}|$ \\ \hline
\textsc{ManifoldE} (hyperplane) & $|(\mathbf{h}+\mathbf{r_{head}})^T(\mathbf{t}+\mathbf{r_{tail}})-D_r|$ & $k|\mathcal{E}|+(2k+1)|\mathcal{R}|$ \\ \hline
\textsc{rescal} & $\mathbf{h}^T\mathbf{W_r}\mathbf{t}$ & $k|\mathcal{E}|+k^2|\mathcal{R}|$ \\ \hline
\textsc{HolE} & $\mathbf{r}^T(\mathbf{h}\star\mathbf{t})$ ($\star$ is circular correlation) & $k|\mathcal{E}|+k|\mathcal{R}|$ \\ \hline
\textsc{ConvE} & $f(vec(f([\bar{\mathbf{h}};\bar{\mathbf{r}}]*\omega))\mathbf{W})\mathbf{t}$ & $k|\mathcal{E}|+k|\mathcal{R}|+kcmn$ \\ \hline
\end{tabular}
\caption{Some selected knowledge graph embedding models. The four models above the double line are considered in this
paper. Except for \textsc{ComplEx}, all boldface lower case letters represent vectors in $\mathbb{R}^k$, and boldface upper case letters represent matrices in $\mathbb{R}^{k\times k}$. $\mathbf{I}$ is the identity matrix.}
\label{tab:models}
\end{table*}

In this work, we provide a generic solution to improve the training of a wide range of knowledge graph embedding models. Inspired by the recent advances of generative adversarial deep models~\cite{goodfellow2014generative}, we propose a novel adversarial learning framework, namely, \textsc{kbgan}, for generating better negative examples to train knowledge graph embedding models. More specifically, we consider probability-based, log-loss embedding models as the generator to supply better quality negative examples, and use distance-based, margin-loss embedding models as the discriminator to generate the final knowledge graph embeddings. Since the generator has a discrete generation step, we cannot directly use the gradient-based approach to backpropagate the errors. We then consider a one-step reinforcement learning setting, and use a variance-reduction \textsc{reinforce} method to achieve this goal. Empirically, we perform experiments on three common KGE datasets (FB15K-237, WN18 and WN18RR), and verify the adversarial learning approach with a set of KGE models. Our experiments show that across various settings, this adversarial learning mechanism can significantly improve the performance of some of the most commonly used translation based KGE methods.
Our contributions are three-fold:
\begin{itemize}
\item We are the first to consider adversarial learning to generate useful negative training examples to improve knowledge graph embedding.
\item This adversarial learning framework applies to a wide range of KGE models, without the need of external ontologies constraints.
\item Our method shows consistent performance gains on three commonly used KGE datasets.
\end{itemize}

\section{Related Work}
\subsection{Knowledge Graph Embeddings}
A large number of knowledge graph embedding models, which represent entities and relations in a knowledge graph with 
vectors or matrices, have been proposed in recent years.  \textsc{rescal}~\cite{nickel2011three} is one of the earliest studies on matrix factorization based knowledge graph embedding models, using a bilinear form as score function.
\textsc{TransE} \cite{bordes2013translating} is the first model to introduce translation-based embedding. Later variants, such as \textsc{TransH} \cite{wang2014knowledge}, \textsc{TransR} \cite{lin2015learning} and \textsc{TransD} \cite{transd}, extend \textsc{TransE} by projecting the embedding vectors of entities into various spaces.
\textsc{DistMult}~\cite{yang2015embedding} simplifies \textsc{rescal} by only using a diagonal matrix, and \textsc{ComplEx} \cite{trouillon2016complex} extends \textsc{DistMult} into the complex number field. \cite{kgreview} is a comprehensive survey on these models.

Some of the more recent models achieve strong performances. \textsc{ManifoldE} \cite{manifolde} embeds a triple as a manifold rather than a point. \textsc{HolE} \cite{hole} employs circular correlation to combine the two entities in a triple. \textsc{ConvE} \cite{conve} uses a convolutional neural network as the score function. However, most of these studies use uniform sampling to generate negative training examples~\cite{bordes2013translating}. Because our framework is independent of the concrete form of models, all these models can be potentially incorporated into our framework, regardless of the complexity. As a proof of principle, our work focuses on simpler models. Table \ref{tab:models} summarizes the score functions and dimensions of all models mentioned above.

\subsection{Generative Adversarial Networks and its Variants}

Generative Adversarial Networks (\textsc{gan}s) \cite{goodfellow2014generative} was originally proposed for generating samples in a continuous space such as images. A \textsc{gan} consists of two parts, the \emph{generator} and the \emph{discriminator}. The generator accepts a noise input and outputs an image. The discriminator is a classifier which classifies images as ``true'' (from the ground truth set) or ``fake'' (generated by the generator). When training a \textsc{gan}, the generator and the discriminator play a minimax game, in which the generator tries to generate ``real'' images to deceive the discriminator, and the discriminator tries to tell them apart from ground truth images. \textsc{gan}s are also capable of generating samples satisfying certain requirements, such as conditional \textsc{gan} \cite{conditionalgan}.

It is not possible to use \textsc{gan}s in its original form for generating discrete samples like natural language sentences or knowledge graph triples, because the discrete sampling step prevents gradients from propagating back to the generator. \textsc{SeqGan} \cite{seqgan} is one of the first successful solutions to this problem by using reinforcement learning---It trains the generator using policy gradient and other tricks. \textsc{irgan} \cite{irgan} is a recent work which combines two categories of information retrieval models into a discrete \textsc{gan} framework. Likewise, our framework relies on policy gradient to train the generator which provides discrete negative triples.

The discriminator in a \textsc{gan} is not necessarily a classifier. Wasserstein \textsc{gan} or \textsc{wgan} \cite{wgan} uses a regressor with clipped parameters as its discriminator, based on solid analysis about the mathematical nature of \textsc{gan}s. \textsc{GoGan} \cite{gogan} further replaces the loss function in \textsc{wgan} with marginal loss. Although originating from very different fields, the form of loss function in our framework turns out to be more closely related to the one in \textsc{GoGan}.

\section{Our Approaches}
In this section, we first define two types of training objectives in knowledge graph embedding models to show how \textsc{kbgan} can be applied. Then, we demonstrate a long overlooked problem about negative sampling which motivates us to propose \textsc{kbgan} to address the problem. Finally, we dive into the mathematical, and algorithmic details of \textsc{kbgan}.

\subsection{Types of Training Objectives}
For a given knowledge graph, let $\mathcal{E}$ be the set of entities, $\mathcal{R}$ be the set of relations, and $\mathcal{T}$ be the set of ground truth triples.
In general, a knowledge graph embedding (KGE) model can be formulated as a \emph{score function} $f(h,r,t), h,t\in\mathcal{E}, r\in \mathcal{R}$ which assigns a score to every possible triple in the knowledge graph. The estimated likelihood of a triple to be true depends only on its score given by the score function.

Different models formulate their score function based on different designs, and therefore interpret scores differently, which further lead to various training objectives. Two common forms of training objectives are particularly of our interest:

\noindent \textbf{Marginal loss function} is commonly used by a large group of models called translation-based models, whose score function models distance between points or vectors, such as \textsc{TransE}, \textsc{TransH}, \textsc{TransR}, \textsc{TransD} and so on. In these models, smaller distance indicates a higher likelihood of truth, but only qualitatively.
The marginal loss function takes the following form:
\begin{equation}
L_{m}=\sum_{(h,r,t)\in\mathcal{T}}[f(h,r,t)-f(h',r,t')+\gamma]_+\label{eq:marginalloss}
\end{equation}
where $\gamma$ is the margin, $[\cdot]_+=\max(0,\cdot)$ is the hinge function, and $(h',r,t')$ is a negative triple. The negative triple is generated by replacing the head entity or the tail entity of a positive triple with a random entity in the knowledge graph, or formally $(h',r,t')\in\{(h',r,t)|h'\in\mathcal{E}\}\cup\{(h,r,t')|t'\in\mathcal{E}\}$.

\begin{figure*}[t]
  \centering
    \includegraphics[width=0.95\textwidth]{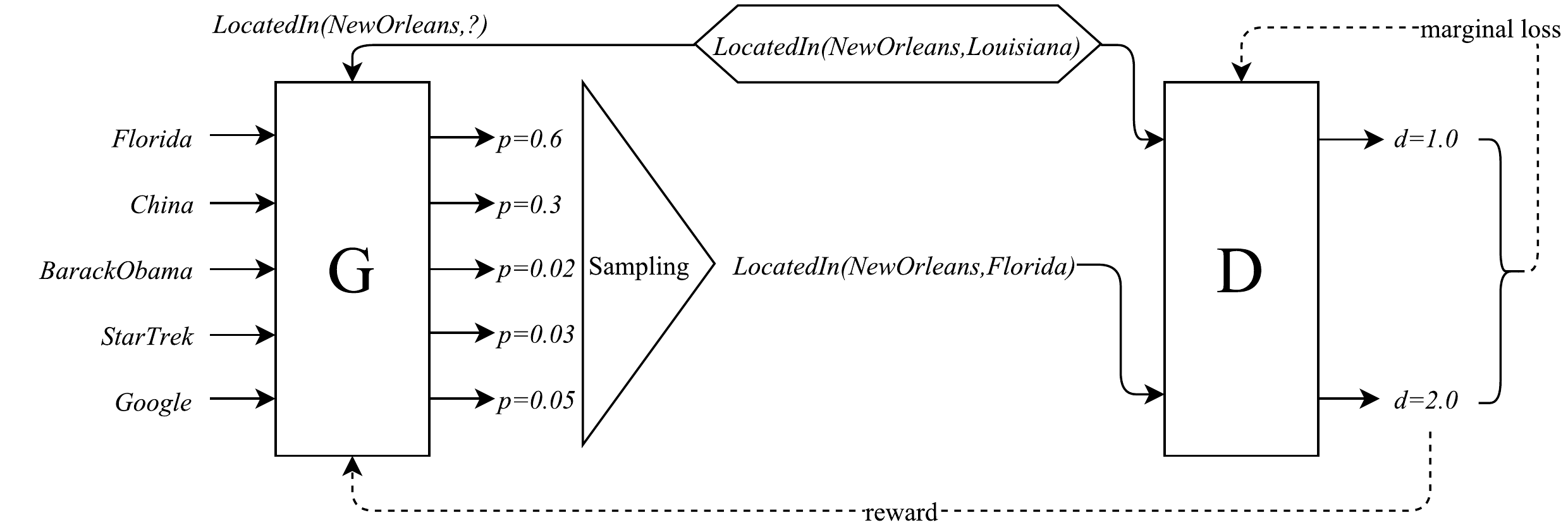}
  \caption{An overview of the \textsc{kbgan} framework. The generator (G) calculates a probability distribution over a set of candidate negative triples, then sample one triples from the distribution as the output. The discriminator (D) receives the generated negative triple as well as the ground truth triple (in the hexagonal box), and calculates their scores. G minimizes the score of the generated negative triple by policy gradient, and D minimizes the marginal loss between positive and negative triples by gradient descent.}
  \label{fig:overview}
\end{figure*}

\noindent \textbf{Log-softmax loss function} is commonly used by models whose score function has probabilistic interpretation. Some notable examples are \textsc{rescal}, \textsc{DistMult}, \textsc{ComplEx}. Applying the softmax function on scores of a given set of triples gives the probability of a triple to be the best one among them: $p(h,r,t)=\frac{\exp f(h,r,t)}{\sum_{(h',r,t')}\exp f(h',r,t')}$. The loss function is the negative log-likelihood of this probabilistic model:
\begin{multline}
L_{l}=\sum_{(h,r,t)\in\mathcal{T}}-\log \frac{\exp f(h,r,t)}{\sum\exp f(h',r,t')}\\
(h',r,t')\in\{(h,r,t)\}\cup Neg(h,r,t)\label{eq:nllloss}
\end{multline}
where $Neg(h,r,t)\subset\{(h',r,t)|h'\in\mathcal{E}\}\cup\{(h,r,t')|t'\in\mathcal{E}\}$ is a set of sampled corrupted triples.

Other forms of loss functions exist, for example \textsc{ConvE} uses a triple-wise logistic function to model how likely the triple is true, but by far the two described above are the most common. Also, softmax function gives an probabilistic distribution over a set of triples, which is necessary for a generator to sample from them.

\subsection{Weakness of Uniform Negative Sampling}

Most previous KGE models use \emph{uniform negative sampling} for generating negative triples, that is, replacing the head or tail entity of a positive triple with any of the entities in $\mathcal{E}$, all with equal probability. Most of the negative triples generated in this way contribute little to learning an effective embedding, because they are too obviously false.

To demonstrate this issue, let us consider the following example. Suppose we have a ground truth triple \emph{LocatedIn(NewOrleans,Louisiana)}, and corrupt it by replacing its tail entity. First, we remove the tail entity, leaving \emph{LocatedIn(NewOrleans,?)}. 
Because the relation \emph{LocatedIn} constraints types of its entities, ``?'' must be a geographical region. If we fill ``?'' with a random entity $e\in\mathcal{E}$, the probability of $e$ having a wrong type is very high, resulting in ridiculous triples like \emph{LocatedIn(NewOrleans,BarackObama)} or \emph{LocatedIn(NewOrleans,StarTrek)}. Such triples are considered ``too easy'', because they can be eliminated solely by types. In contrast, \emph{LocatedIn(NewOrleans,Florida)} is a very useful negative triple, because it satisfies type constraints, but it cannot be proved wrong without detailed knowledge of American geography. If a KGE model is fed with mostly ``too easy'' negative examples, it would probably only learn to represent types, not the underlying semantics.

The problem is less severe to models using log-softmax loss function, because they typically samples tens or hundreds of negative triples for one positive triple in each iteration, and it is likely to have a few useful negatives among them. For instance, \cite{trouillon2016complex} found that a 100:1 negative-to-positive ratio results in the best performance for \textsc{ComplEx}. However, for marginal loss function, whose negative-to-positive ratio is always 1:1, the low quality of uniformly sampled negatives can seriously damage their performance.

\begin{algorithm*}[t]
\small
 \KwData{training set of positive fact triples $\mathcal{T}=\{(h,r,t)\}$}
 \KwIn{Pre-trained generator G with parameters $\theta_G$ and score function $f_G(h,r,t)$, and pre-trained discriminator D with parameters $\theta_D$ and score function $f_D(h,r,t)$}
 \KwOut{Adversarially trained discriminator}
 $b \longleftarrow 0$\tcp*[l]{baseline for policy gradient}
 \Repeat{convergence}{
  Sample a mini-batch of data $\mathcal{T}_{batch}$ from $\mathcal{T}$ \;
  $G_G \longleftarrow 0$, $G_D \longleftarrow 0$\tcp*[l]{gradients of parameters of G and D}
  $r_{sum} \longleftarrow 0$\tcp*[l]{for calculating the baseline}
  \For{$(h,r,t)\in \mathcal{T}_{batch}$}{
    Uniformly randomly sample $N_s$ negative triples $Neg(h,r,t)=\{(h_i',r,t_i')\}_{i=1\dots N_s}$\;
    Obtain their probability of being generated: $p_i=\frac{\exp f_G(h_i', r, t_i')}{\sum_{j=1}^{N_s} \exp f_G(h_j',r,t_j')}$\;
    Sample one negative triple $(h_s',r,t_s')$ from $Neg(h,r,t)$ according to $\{p_i\}_{i=1\dots N_s}$. Assume its probability to be $p_s$\;
    $G_D \longleftarrow G_D + \nabla_{\theta_D}[f_D(h,r,t)-f_D(h_s',r,t_s')+\gamma]_+$\tcp*[l]{accumulate gradients for D}
    $r \longleftarrow -f_D(h_s',r,t_s'), r_{sum} \longleftarrow r_{sum} + r$\tcp*[l]{$r$ is the reward}
    $G_G \longleftarrow G_G + (r-b)\nabla_{\theta_G} \log p_s$\tcp*[l]{accumulate gradients for G}
  }
  $\theta_G \longleftarrow \theta_G+\eta_G G_G, \theta_D \longleftarrow \theta_D-\eta_D G_D$\tcp*[l]{update parameters}
  $b \leftarrow r_{sum} / |\mathcal{T}_{batch}|$\tcp*[l]{update baseline}
 }
 \caption{The \textsc{kbgan} algorithm}
 \label{alg:kbgan}
\end{algorithm*}

\subsection{Generative Adversarial Training for\\ Knowledge Graph Embedding Models}

Inspired by GANs, we propose an adversarial training framework named \textsc{kbgan} which uses a KGE model with \textbf{softmax probabilities} to provide high-quality negative samples for the training of a KGE model whose training objective is \textbf{marginal loss function}. This framework is independent of the score functions of these two models, and therefore possesses some extent of universality. Figure \ref{fig:overview} illustrates the overall structure of \textsc{kbgan}.

In parallel to terminologies used in \textsc{gan} literature, we will simply call these two models \emph{generator} and \emph{discriminator} respectively in the rest of this paper. We use softmax probabilistic models as the generator because they can adequately model the ``sampling from a probability distribution'' process of discrete \textsc{gan}s, and we aim at improving discriminators based on marginal loss because they can benefit more from high-quality negative samples. Note that a major difference between \textsc{gan} and our work is that, the ultimate goal of our framework is to produce a good discriminator, whereas \textsc{gans} are aimed at training a good generator. In addition, the discriminator here is not a classifier as it would be in most \textsc{gan}s. 

Intuitively, the discriminator should assign a relatively small distance to a high-quality negative sample. In order to encourage the generator to generate useful negative samples, the objective of the generator is to minimize the distance given by discriminator for its generated triples. And just like the ordinary training process, the objective of the discriminator is to minimize the marginal loss between the positive triple and the generated negative triple. In an adversarial training setting, the generator and the discriminator are alternatively trained towards their respective objectives.

Suppose that the generator produces a probability distribution on negative triples $p_G(h',r,t'|h,r,t)$ given a positive triple $(h,r,t)$, and generates negative triples $(h',r,t')$ by sampling from this distribution. Let $f_D(h,r,t)$ be the score function of the discriminator. The objective of the discriminator can be formulated as minimizing the following marginal loss function:
\begin{multline}
L_D=\sum_{(h,r,t)\in\mathcal{T}}[f_D(h,r,t)-f_D(h',r,t')+\gamma]_+ \\
(h',r,t')\sim p_G(h',r,t'|h,r,t)
\end{multline}
The only difference between this loss function and Equation \ref{eq:marginalloss} is that it uses negative samples from the generator.

\begin{table*}[t]
\centering
\begin{tabular}{|c|c|c|}
\hline
\textbf{Model} & \textbf{Hyperparameters} & \textbf{Constraints or Regularizations}  \\ \hline
\textsc{TransE} & $L_1$ distance, $k=50, \gamma=3$ & $||\mathbf{e}||_2\leq 1,||\mathbf{r}||_2\leq 1$ \\ \hline
\textsc{TransD} & $L_1$ distance, $k=50, \gamma=3$ & $||\mathbf{e}||_2\leq 1,||\mathbf{r}||_2\leq 1,||\mathbf{e_p}||_2\leq 1,||\mathbf{r_p}||_2\leq 1$  \\ \hline
\textsc{DistMult} & $k=50, \lambda=1/0.1$ & L2 regularization: $L_{reg}=L+\lambda||\Theta||_2^2$ \\ \hline
\textsc{ComplEx}  & $2k=50, \lambda=1/0.1$ & L2 regularization: $L_{reg}=L+\lambda||\Theta||_2^2$ \\ \hline
\end{tabular}
\caption{Hyperparameter settings of the 4 models we used. For \textsc{DistMult} and \textsc{ComplEx}, $\lambda=1$ is used for FB15k-237 and $\lambda=0.1$ is used for WN18 and WN18RR. All other hyperparameters are shared among all datasets. $L$ is the global loss defined in Equation \eqref{eq:nllloss}. $\Theta$ represents all parameters in the model.}
\label{tab:hyperparams}
\end{table*}

\begin{table}[t]
\small
\centering
\begin{tabular}{|l|l|l|l|l|l|}
\hline
\textbf{Dataset} & \textbf{\#r} & \textbf{\#ent.} & \textbf{\#train} & \textbf{\#val} & \textbf{\#test} \\ \hline
FB15k-237 & 237 & 14,541 & 272,115 & 17,535 & 20,466 \\ \hline
WN18 & 18 & 40,943 & 141,442 & 5,000 & 5,000 \\ \hline
WN18RR & 11 & 40,943 & 86,835 & 3,034 & 3,134 \\ \hline
\end{tabular}
\caption{Statistics of datasets we used in the experiments. ``r'': relations.}
\label{tab:datasets}
\end{table}

The objective of the generator can be formulated as maximizing the following expectation of negative distances:
\begin{multline}
R_G=\sum_{(h,r,t)\in\mathcal{T}}\mathbb{E}[-f_D(h',r,t')] \\
(h',r,t')\sim p_G(h',r,t'|h,r,t)
\end{multline}

$R_G$ involves a discrete sampling step, so we cannot find its gradient with simple differentiation. We use a simple special case of Policy Gradient Theorem\footnote{A proof can be found in the supplementary material} \cite{policygradient} to obtain the gradient of $R_G$ with respect to parameters of the generator:
\begin{multline}
\nabla_G R_G=\sum_{(h,r,t)\in\mathcal{T}}\mathbb{E}_{(h',r,t')\sim p_G(h',r,t'|h,r,t)} \\
[-f_D(h',r,t')\nabla_G \log p_G(h',r,t'|h,r,t)] \\
\simeq \sum_{(h,r,t)\in\mathcal{T}}\frac{1}{N}\sum_{(h_i',r,t_i')\sim p_G(h',r,t'|h,r,t), i=1\dots N} \\
[-f_D(h',r,t')\nabla_G \log p_G(h',r,t'|h,r,t)]
\end{multline}
where the second approximate equality means we approximate the expectation with sampling in practice. Now we can calculate the gradient of $R_G$ and optimize it with gradient-based algorithms.

Policy Gradient Theorem arises from reinforcement learning (RL), so we would like to draw an analogy between our model and an RL model. The generator can be viewed as an \emph{agent} which interacts with the \emph{environment} by performing \emph{actions} and improves itself by maximizing the \emph{reward} returned from the environment in response of its actions. Correspondingly, the discriminator can be viewed as the environment. Using RL terminologies, $(h,r,t)$ is the \emph{state} (which determines what actions the actor can take), $p_G(h',r,t'|h,r,t)$ is the \emph{policy} (how the actor choose actions), $(h',r,t')$ is the \emph{action}, and $-f_D(h',r,t')$ is the \emph{reward}. The method of optimizing $R_G$ described above is called \textsc{reinforce}~\cite{williams1992simple} algorithm in RL. Our model is a simple special case of RL, called one-step RL. In a typical RL setting, each action performed by the agent will change its state, and the agent will perform a series of actions (called an \emph{epoch}) until it reaches certain states or the number of actions reaches a certain limit. However, in the analogy above, actions does not affect the state, and after each action we restart with another unrelated state, so each epoch consists of only one action.

To reduce the variance of \textsc{reinforce} algorithm, it is common to subtract a \emph{baseline} from the reward, which is an arbitrary number that only depends on the state, without affecting the expectation of gradients.\footnote{A proof of such fact can also be found in the supplementary material} In our case, we replace $-f_D(h',r,t')$ with $-f_D(h',r,t')-b(h,r,t)$ in the equation above to introduce the baseline. To avoid introducing new parameters, we simply let $b$ be a constant, the average reward of the whole training set: $b=\sum_{(h,r,t)\in\mathcal{T}}\mathbb{E}_{(h',r,t')\sim p_G(h',r,t'|h,r,t)}[-f_D(h',r,t')]$. In practice, $b$ is approximated by the mean of rewards of recently generated negative triples.

Let the generator's score function to be $f_G(h,r,t)$, given a set of candidate negative triples $Neg(h,r,t)\subset\{(h',r,t)|h'\in\mathcal{E}\}\cup\{(h,r,t')|t'\in\mathcal{E}\}$, the probability distribution $p_G$ is modeled as:
\begin{multline}
p_G(h',r,t'|h,r,t)=\frac{\exp f_G(h',r,t')}{\sum\exp f_G(h^*,r,t^*)} \\
(h^*,r,t^*)\in Neg(h,r,t)
\end{multline}

Ideally, $Neg(h,r,t)$ should contain all possible negatives. However, knowledge graphs are usually highly incomplete, so the "hardest" negative triples are very likely to be false negatives (true facts). To address this issue, we instead generate $Neg(h,r,t)$ by uniformly sampling of $N_s$ entities (a small number compared to the number of all possible negatives) from $\mathcal{E}$ to replace $h$ or $t$. Because in real-world knowledge graphs, true negatives are usually far more than false negatives, such set would be unlikely to contain any false negative, and the negative selected by the generator would likely be a true negative. Using a small $Neg(h,r,t)$ can also significantly reduce computational complexity.

Besides, we adopt the ``bern'' sampling technique \cite{wang2014knowledge} which replaces the ``1'' side in ``1-to-N'' and ``N-to-1'' relations with higher probability to further reduce false negatives.

Algorithm \ref{alg:kbgan} summarizes the whole adversarial training process. Both the generator and the discriminator require pre-training, which is the same as conventionally training a single KBE model with uniform negative sampling. Formally speaking, one can pre-train the generator by minimizing the loss function defined in Equation \eqref{eq:marginalloss}, and pre-train the discriminator by minimizing the loss function defined in Equation \eqref{eq:nllloss}. Line 14 in the algorithm assumes that we are using the vanilla gradient descent as the optimization method, but obviously one can substitute it with any gradient-based optimization algorithm.

\begin{table*}[t]
\centering
\begin{tabular}{|l|cc|cc|cc|}
\hline
             & \multicolumn{2}{c}{\bf FB15k-237} \vline & \multicolumn{2}{c}{\bf WN18} \vline & \multicolumn{2}{c}{\bf WN18RR}\vline \\
Method                         & MRR         & H@10   & MRR         & H@10 & MRR & H@10\\
\hline
\textsc{TransE}    & - & 42.8$^{\dag}$ & - & 89.2 & -    & 43.2$^{\dag}$ \\
\textsc{TransD}    & - & 45.3$^{\dag}$ & - & 92.2 & -    & 42.8$^{\dag}$ \\ 
\textsc{DistMult}  & 24.1$^{\ddag}$  & 41.9$^{\ddag}$ & 82.2 & 93.6 & 42.5$^{\ddag}$ & 49.1$^{\ddag}$ \\
\textsc{ComplEx}   & 24.0$^{\ddag}$  & 41.9$^{\ddag}$ & \textbf{94.1} & 94.7 & \textbf{44.4}$^{\ddag}$ & \textbf{50.7}$^{\ddag}$ \\
\hline
\textsc{TransE} (pre-trained)                         & 24.2    & 42.2  & 43.3   & 91.5 & 18.6 & 45.9 \\
\textsc{kbgan} (\textsc{TransE} + \textsc{DistMult})  & 27.4  & 45.0 & 71.0 & \textbf{\underline{94.9}} & 21.3 & \underline{48.1} \\
\textsc{kbgan} (\textsc{TransE} + \textsc{ComplEx})   & \textbf{\underline{27.8}} & 45.3 & 70.5  & \textbf{\underline{94.9}} & 21.0 & 47.9 \\
\textsc{TransD} (pre-trained)                         & 24.5 & 42.7 & 49.4  & 92.8 & 19.2 & 46.5 \\
\textsc{kbgan} (\textsc{TransD} + \textsc{DistMult})  & \textbf{\underline{27.8}} & \textbf{\underline{45.8}} & 77.2 & 94.8 & 21.4 & 47.2\\
\textsc{kbgan} (\textsc{TransD} + \textsc{ComplEx})   & 27.7 & \textbf{\underline{45.8}} & \underline{77.9} & 94.8 & \underline{21.5} & 46.9\\ \hline
\end{tabular}
\caption{Experimental results. Results of \textsc{kbgan} are results of its discriminator (on the left of the ``+'' sign). Underlined results are the best ones among our implementations. Results marked with $\dag$ are produced by running Fast-TransX \cite{lin2015learning} with its default parameters. Results marked with $\ddag$ are copied from \cite{conve}. All other baseline results are copied from their original papers.}
\label{tab:results}
\end{table*}

\section{Experiments}

To evaluate our proposed framework, we test its performance for the link prediction task with different generators and discriminators. For the generator, we choose two classical probability-based KGE model, \textsc{DistMult} and \textsc{ComplEx}, and for the discriminator,  we also choose two classical translation-based KGE model, \textsc{TransE} and \textsc{TransD}, resulting in four possible combinations of generator and discriminator in total. See Table \ref{tab:models} for a brief summary of these models.

\subsection{Experimental Settings}
\subsubsection{Datasets}
We use three common knowledge base completion datasets for our experiment: FB15k-237, WN18 and WN18RR. FB15k-237 is a subset of FB15k introduced by \cite{Toutanova2015}, which removed redundant relations in FB15k and greatly reduced the number of relations. Likewise, WN18RR is a subset of WN18 introduced by \cite{conve} which removes reversing relations and dramatically increases the difficulty of reasoning. Both FB15k and WN18 are first introduced by \cite{bordes2013translating} and have been commonly used in knowledge graph researches. Statistics of datasets we used are shown in Table \ref{tab:datasets}.

\subsubsection{Evaluation Protocols}
Following previous works like \cite{yang2015embedding} and \cite{trouillon2016complex}, for each run, we report two common metrics, mean reciprocal ranking (MRR) and hits at 10 (H@10). We only report scores under the \emph{filtered} setting \cite{bordes2013translating}, which removes all triples appeared in training, validating, and testing sets from candidate triples before obtaining the rank of the ground truth triple.

\begin{figure*}[!t]
\centering
\subfloat{\includegraphics[width=0.333\textwidth]{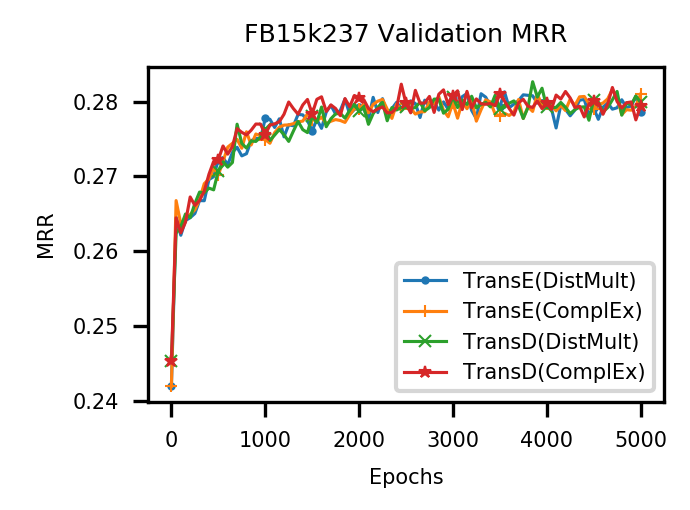}}
\subfloat{\includegraphics[width=0.333\textwidth]{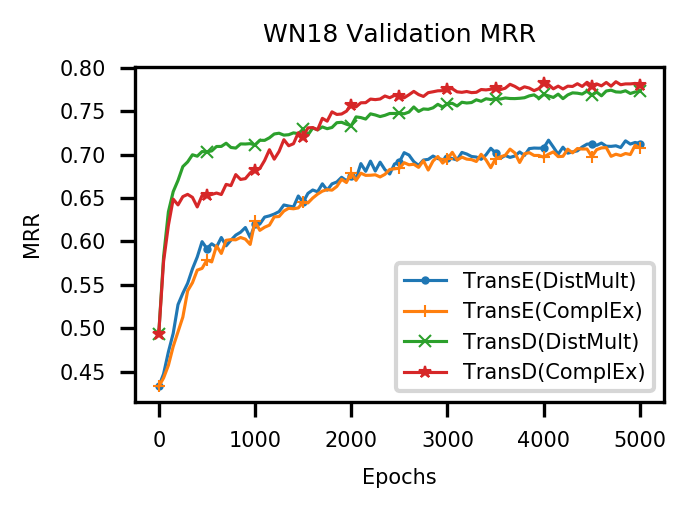}}
\subfloat{\includegraphics[width=0.333\textwidth]{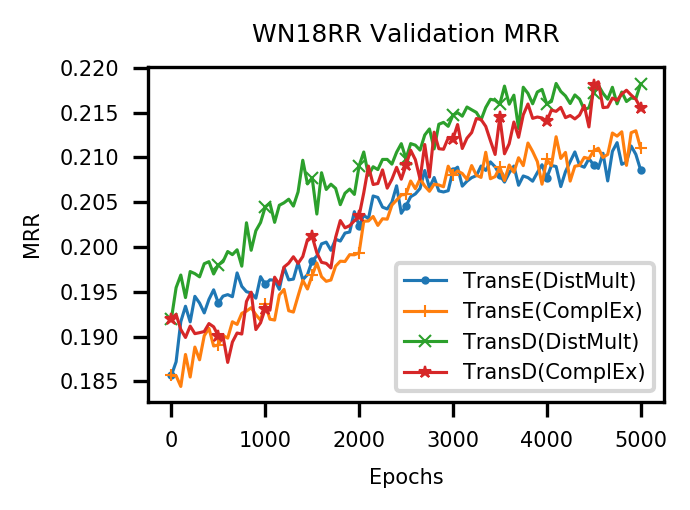}}\\
\subfloat{\includegraphics[width=0.333\textwidth]
{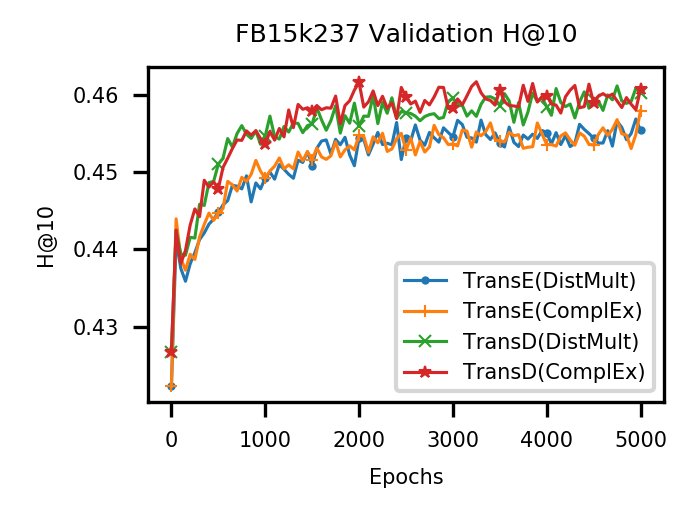}}
\subfloat{\includegraphics[width=0.333\textwidth]{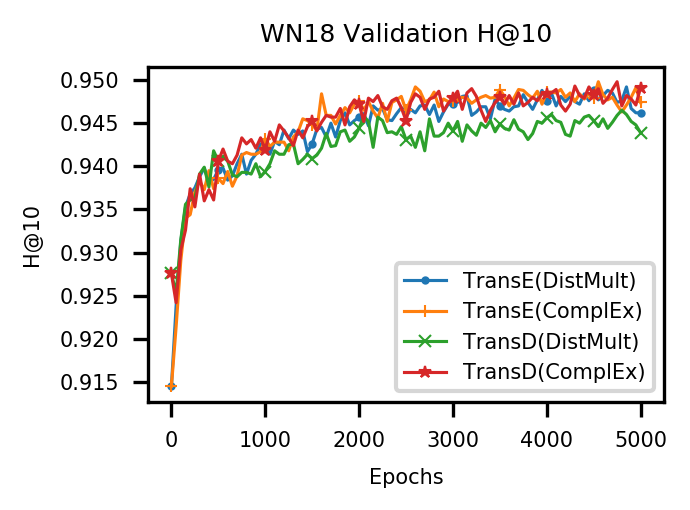}}
\subfloat{\includegraphics[width=0.333\textwidth]{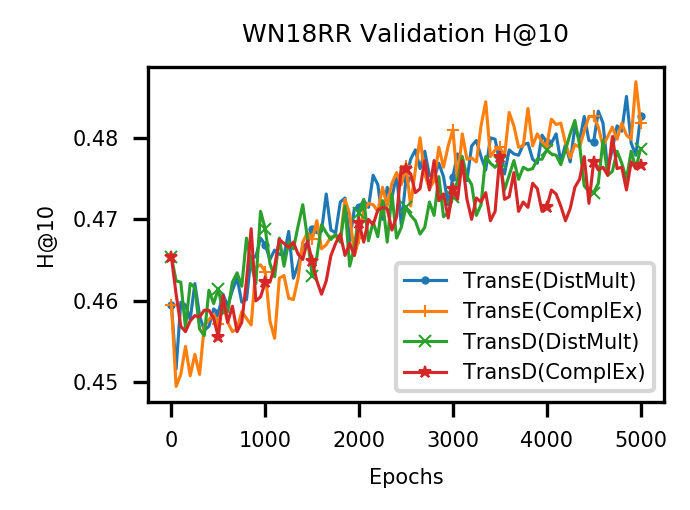}}
\caption{Learning curves of \textsc{kbgan}. All metrics improve steadily as training proceeds.}
\label{fig:curves} 
\end{figure*}

\subsubsection{Implementation Details}\footnote{The \textsc{kbgan} source code is available at \url{https://github.com/cai-lw/KBGAN}}
In the pre-training stage, we train every model to convergence for 1000 epochs, and divide every epoch into 100 mini-batches. To avoid overfitting, we adopt early stopping by evaluating MRR on the validation set every 50 epochs. We tried $\gamma=0.5,1,2,3,4,5$ and $L_1, L_2$ distances for \textsc{TransE} and \textsc{TransD}, and $\lambda=0.01,0.1,1,10$ for \textsc{DistMult} and \textsc{ComplEx}, and determined the best hyperparameters listed on table \ref{tab:hyperparams}, based on their performances on the validation set after pre-training. Due to limited computation resources, we deliberately limit the dimensions of embeddings to $k=50$, similar to the one used in earlier works, to save time. We also apply certain constraints or regularizations to these models, which are mostly the same as those described in their original publications, and also listed on table \ref{tab:hyperparams}.

In the adversarial training stage, we keep all the hyperparamters determined in the pre-training stage unchanged. The number of candidate negative triples, $N_s$, is set to 20 in all cases, which is proven to be optimal among the candidate set of $\{5, 10, 20, 30, 50\}$. We train for 5000 epochs, with 100 mini-batches for each epoch. We also use early stopping in adversarial training by evaluating MRR on the validation set every 100 epochs.

We use the self-adaptive optimization method Adam \cite{adam} for all trainings, and always use the recommended default setting $\alpha=0.001, \beta_1=0.9, \beta_2=0.999, \epsilon=10^{-8}$.

\subsection{Results}
Results of our experiments as well as baselines are shown in Table \ref{tab:results}. All settings of adversarial training bring a pronounced improvement to the model, which indicates that our method is consistently effective in various cases. \textsc{TransE} performs slightly worse than \textsc{TransD} on FB15k-237 and WN18, but better on WN18RR. Using \textsc{DistMult} or \textsc{ComplEx} as the generator does not affect performance greatly.

\textsc{TransE} and \textsc{TransD} enhanced by \textsc{kbgan} can significantly beat their corresponding baseline implementations, and outperform stronger baselines in some cases. As a prototypical and proof-of-principle experiment, we have never expected state-of-the-art results. Being simple models proposed several years ago, \textsc{TransE} and \textsc{TransD} has their limitations in expressiveness that are unlikely to be fully compensated by better training technique. In future researches, people may try employing more advanced models into \textsc{kbgan},  and we believe it has the potential to become state-of-the-art.

To illustrate our training progress, we plot performances of the discriminator on validation set over epochs, which are displayed in Figure \ref{fig:curves}. As all these graphs show, our performances are always in increasing trends, converging to its maximum as training proceeds, which indicates that \textsc{kbgan} is a robust \textsc{gan} that can converge to good results in various settings, although \textsc{gan}s are well-known for difficulty in convergence. Fluctuations in these graphs may seem more prominent than other KGE models, but is considered normal for an adversially trained model. Note that in some cases the curve still tends to rise after 5000 epochs. We do not have sufficient computation resource to train for more epochs, but we believe that they will also eventually converge.



\begin{table*}[t]
\centering
\begin{tabular}{|p{48mm}|p{48mm}|p{48mm}|}
\hline
\textbf{Positive fact} & \textbf{Uniform random sample} & \textbf{Trained generator}  \\ \hline
(condensation\_NN\_2, \newline derivationally\_related\_form, \newline \textbf{distill\_VB\_4}) & family\_arcidae\_NN\_1 \newline repast\_NN\_1 \newline beater\_NN\_2 \newline coverall\_NN\_1 \newline cash\_advance\_NN\_1 & revivification\_NN\_1 \newline mouthpiece\_NN\_3 \newline \textit{liquid\_body\_substance\_NN\_1} \newline stiffen\_VB\_2 \newline \textit{hot\_up\_VB\_1} \\ \hline
(\textbf{colorado\_river\_NN\_2}, \newline instance\_hypernym, \newline river\_NN\_1) & lunar\_calendar\_NN\_1 \newline umbellularia\_californica\_NN\_1 \newline tonality\_NN\_1 \newline creepy-crawly\_NN\_1 \newline moor\_VB\_3 & \textit{idaho\_NN\_1} \newline \textit{sayan\_mountains\_NN\_1} \newline \textit{lower\_saxony\_NN\_1} \newline order\_ciconiiformes\_NN\_1 \newline jab\_NN\_3  \\ \hline
(\textbf{meeting\_NN\_2}, \newline hypernym, \newline social\_gathering\_NN\_1) & cellular\_JJ\_1 \newline commercial\_activity\_NN\_1 \newline giant\_cane\_NN\_1 \newline streptomyces\_NN\_1 \newline tranquillize\_VB\_1 & \textit{attach\_VB\_1} \newline \textit{bond\_NN\_6} \newline heavy\_spar\_NN\_1 \newline satellite\_NN\_1 \newline peep\_VB\_3 \\ \hline
\end{tabular}
\caption{Examples of negative samples in WN18 dataset. The first column is the positive fact, and the term in bold is the one to be replaced by an entity in the next two columns. The second column consists of random entities drawn from the whole dataset. The third column contains negative samples generated by the generator in the last 5 epochs of training. Entities in italic are considered to have semantic relation to the positive one}
\label{tab:examples}
\end{table*}

\subsection{Case study}

To demonstrate that our approach does generate better negative samples, we list some examples of them in Table \ref{tab:examples}, using the \textsc{kbgan} (\textsc{TransE} + \textsc{DistMult}) model and the WN18 dataset. All hyperparameters are the same as those described in Section 4.1.3.

Compared to uniform random negatives which are almost always totally unrelated, the generator generates more \emph{semantically} related negative samples, which is different from type relatedness we used as example in Section 3.2, but also helps training. In the first example, two of the five terms are physically related to the process of distilling liquids. In the second example, three of the five entities are geographical objects. In the third example, two of the five entities express the concept of ``gather''.

Because we deliberately limited the strength of generated negatives by using a small $N_s$ as described in Section 3.3, the semantic relation is pretty weak, and there are still many unrelated entities. However, empirical results (when selecting the optimal $N_s$) shows that such situation is more beneficial for training the discriminator than generating even stronger negatives.

\section{Conclusions}
We propose a novel adversarial learning method for improving a wide range of knowledge graph embedding models---We designed a generator-discriminator framework with dual KGE components. Unlike random uniform sampling, the generator model generates higher quality negative examples, which allow the discriminator model to learn better. To enable backpropagation of error, we introduced a one-step \textsc{reinforce} method to seamlessly integrate the two modules. Experimentally, we tested the proposed ideas with four commonly used KGE models on three datasets, and the results showed that the adversarial learning framework brought consistent improvements to various KGE models under different settings.

\newpage
\bibliography{all.bib}
\bibliographystyle{acl_natbib}

\end{document}